\documentclass[runningheads]{llncs}
\usepackage{graphicx}
\usepackage{acronym}
\usepackage{microtype}
\usepackage{pifont}
\usepackage{tabularx}
\usepackage{multirow}

\usepackage{hyperref}
\usepackage{amsmath}

\usepackage{rotating}
\usepackage{pdflscape}
\usepackage{tablefootnote}

\usepackage[T1]{fontenc}
\usepackage{lmodern}

\newcommand{\cmark}{\ding{51}}%
\newcommand{\xmark}{\ding{55}}
\newcommand{\tbf}{\textbf}

\acrodef{BMBF}{German Federal Ministry of Education and Research}
\acrodef{DL}{Deep Learning}
\acrodef{DoS}{Denial-of-Service}
\acrodef{DVFS}{dynamic voltage and frequency scaling}
\acrodef{EE}{Early Exit}
\acrodef{EENN}{Early Exit Neural Network}
\acrodef{ECG}{Electrocardiography}

\acrodef{GPU}{graphics processing unit}

\acrodef{HPC}{high-performance computing}
\acrodef{IoT}{Internet of Things}
\acrodef{IFM}{intermediate feature map}

\acrodef{MI}{myocardial infarction}
\acrodef{MAC}{Multiply-Accumulate}
\acrodef{MCU}{microcontroller}
\acrodef{NAS}{Network Architecture Search}
\acrodef{NA}{Network Augmentation}
\acrodef{NN}{Neural Network}

\acrodef{p.p.}{percentage point}
\acrodef{RNN}{Recurrent Neural Network}

\acrodef{SoC}{System-on-Chip}

\acrodef{WSL}{Windows Subsystem for Linux}
\acrodef{VM}{Virtual Machine}

\usepackage{array}
\newcolumntype{C}[1]{>{\centering\arraybackslash}p{#1}}
\newcolumntype{Y}{>{\centering\arraybackslash}X}

\begin{document}
\title{Efficient Post-Training Augmentation for Adaptive Inference in Heterogeneous and Distributed IoT Environments 
}
\titlerunning{Network Augmentation for Distributed EENN in IoT Environments}
 %
 %
 \author{Max Sponner\inst{1, 4}\orcidID{0000-0002-4830-9440} \and
 Lorenzo Servadei\inst{2}\orcidID{0000-0003-4322-834X} \and
 Bernd Waschneck\inst{3}\orcidID{0000-0003-0294-8594} \and
 Robert Wille\inst{2}\orcidID{0000-0002-4993-7860} \and 
 Akash Kumar \inst{4}\orcidID{0000-0002-4830-9440}}
 \authorrunning{M. Sponner et al.}
 %
 \institute{Infineon Technologies Dresden GmbH \& Co. KG, Dresden 01099, Germany \and
 Chair for Design Automation - TU Munich, Munich 80333, Germany \and
 Infineon Technologies AG, Neubiberg 85579, Germany \and
 Chair of Processor Design, CfAED - TU Dresden, Dresden 01069, Germany}

\maketitle              
\begin{abstract}
Early Exit Neural Networks (EENNs) present a solution to enhance the efficiency of neural network deployments.
However, creating EENNs is challenging and requires specialized domain knowledge, due to the large amount of additional design choices.
To address this issue, we propose an automated augmentation flow that focuses on converting an existing model into an EENN.
It performs all required design decisions for the deployment to heterogeneous or distributed hardware targets:
Our framework constructs the EENN architecture, maps its subgraphs to the hardware targets, and configures its decision mechanism. 
To the best of our knowledge, it is the first framework that is able to perform all of these steps.

We evaluated our approach on a collection of Internet-of-Things and standard image classification use cases.
For a speech command detection task, our solution was able to reduce the mean operations per inference by 59.67\%. 
For an ECG classification task, it was able to terminate all samples early, reducing the mean inference energy by 74.9\% and computations by 78.3\%. 
On CIFAR-10, our solution was able to achieve up to a 58.75\% reduction in computations.

The search on a ResNet-152 base model for CIFAR-10 took less than nine hours on a laptop CPU.
The low search cost improves the accessibility of EENNs, with the potential to improve the efficiency of neural networks in a wide range of practical applications.

\end{abstract}

\section{Introduction}

\Acp{EENN} are a possible solution to reduce the mean inference cost of \acp{NN}.
This approach involves inserting additional classifier branches between the network's hidden layers that perform the same task as the original classifier.
By dynamically terminating the inference at one of these \acp{EE}, the computational cost and latency of the inference can be reduced.
This makes them ideal for improving the energy efficiency of \ac{DL} applications.
However, designing and implementing an \ac{EENN} for a specific scenario requires expertise in configuring the network architecture and at-runtime termination decision mechanism.
An additional limitation of the design process is that an incorrect configuration can significantly increase inference costs or significantly degrade the prediction quality.

\ac{NAS} frameworks can automate these tasks.
Yet, they often rely on supernets or multi-tiered evolutionary algorithms, resulting in long search times and a large demand for compute resources. 
This makes such frameworks inaccessible for many developers who do not have the necessary resources. 

Instead of performing \ac{NAS} starting from a costly supernet, we aim to enable developers to utilize \acp{EENN} when deploying their already trained traditional \acp{NN}.
Our proposed flow performs \ac{NA}: it converts an existing model - designed and trained by an expert - into an \ac{EENN}, maps its subgraphs to a heterogeneous or distributed device, and configures its decision mechanism.
The flow performs all these required deployment steps automatically and in a efficient way to enable its usage on standard consumer-grade hardware.
The goal is to make the benefits of \acp{EENN} accessible to more developers, by automating the complicated design steps while also maintaining a low cost for the conversion process.

Our framework is designed for \ac{IoT} scenarios, with a focus on its accessibility for developers who do not have access to the compute capabilities of \ac{HPC} clusters.
To the best of our knowledge, our framework is the first to offer this range of functionality. 

The paper is organized as follows:
we review related work on \acp{EENN} in the Related Work section, describe the proposed \ac{NA} framework in the Methodology section, present the results and analysis of the framework's performance in the Evaluation section, and finally, summarize our findings and discuss the implications of our work in the Conclusion section.

\section{Related Work}

\Aclp{EENN} (\acsp{EENN}) and automated \ac{NAS} for \acp{EENN} have been the subject of a growing body of research in recent years.

\subsection{Early Exit Neural Networks}

\begin{figure}
    \centering
    \includegraphics[width=0.9\linewidth]{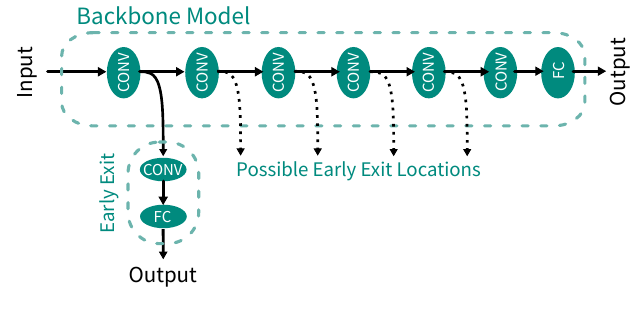}
    \caption{An \ac{EENN} architecture, which combines a backbone model with an \ac{EE}. The considered but not utilized \ac{EE} locations have been marked in the network graph.}
    \label{fig:eenn}
\end{figure}

\Acp{EENN} are a class of \acp{NN} that adapt the inference process at runtime based on the current circumstances.
They are an extension of the concept of Big/Little neural networks, which combine a small and a larger model~\cite{parkBigLittleDeep2015b}.
The underlying idea is to utilize the larger model only if the small model is sufficient for the current sample.
\Acp{EENN} extend this approach by sharing backbone layers between these models and enabling for more than two models to be incorporated.
The overall concept was introduced with BranchyNet~\cite{teerapittayanonBranchyNetFastInference2016a} and has been visualized in Fig.~\ref{fig:eenn}.

An at-runtime decision mechanism performs the selection between classifiers and can be implemented in various ways.
A common solution is to utilize the confidence of the already evaluated classifiers to decide on the current inference's termination~\cite{teerapittayanonBranchyNetFastInference2016a}.
Other approaches include considering the available resource budget~\cite{huLearningAnytimePredictions2018}, 
or using an additional agent model that was trained to perform this decision~\cite{odenaChangingModelBehavior2017}. 

\Acp{EENN} provide several advantages over traditional \acp{NN}, including faster inference times, reduced energy consumption, and improved accuracy on certain tasks.
However, designing effective decision mechanisms and finding optimal architectures and locations for the \acp{EE} that balance accuracy and efficiency remains a challenge.

\subsection{Network Architecture Search for Early Exit Neural Networks}

Designing an \ac{EENN} requires the consideration of additional hyperparameters and design decisions.
These include the locations, count, and architectures of the \acp{EE}; the decision mechanism and its thresholds; as well as the training strategy.
To assist developers in this, \ac{NAS} solutions automate these configurations.
The range of covered functionalities varies between implementations, with some frameworks focused solely on the optimal branch location, which is already an NP-complete problem~\cite{chiangOptimalBranchLocation2021}.
While others include the calibration of the output confidence score~\cite{odemaEExNASEarlyExitNeural2021}, mapping to heterogeneous platforms~\cite{bouzidiMapandConquerEnergyEfficientMapping2023b}, or the incorporation of additional optimizations~\cite{bouzidiMapandConquerEnergyEfficientMapping2023b,bouzidiHADASHardwareAwareDynamic2023}.
Table~\ref{tab:compare_nas} provides a comparison of our novel contribution in terms of features covered by other solutions.
Our tool is - to the best of our knowledge - the first framework that also performs the configuration of exit-wise decision thresholds during the process.

Current solutions primarily utilize genetic search algorithms~\cite{bouzidiMapandConquerEnergyEfficientMapping2023b,bouzidiHADASHardwareAwareDynamic2023,gambellaEDANASAdaptiveNeural2023a}, although multi-objective bayesian search~\cite{odemaEExNASEarlyExitNeural2021} and dynamic programming~\cite{chiangOptimalBranchLocation2021} have also been used.
However, a key limitation of these solutions is the high search cost. 

We want to implement our \ac{NA} flow as a solution for data scientists and engineers that have already developed and trained a model architecture that suits their needs in terms of predictive performance and want to deploy it to a heterogeneous or distributed embedded/\ac{IoT} environment while improving the efficiency by utilizing \acp{EENN}.
This comes with different inputs, requirements and goals than the state-of-the-art \ac{NAS} solutions, which is the reason why we named or approach \ac{NA} as it augments an existing model with additional branches and the decision mechanism to select the most suitable classifier at runtime.

Our conversion tool appears to be the first automated flow that incorporates the ability to configure exit-wise confidence thresholds for the at-runtime decision as part of its functionality.
Our publication focuses on addressing the high search cost by improving its speed and efficiency through reuse and on making \acp{EENN} more accessible by enabling developers to easily convert their existing standard models after training.

\begin{table}
    \centering
    \footnotesize
    \resizebox{\linewidth}{!}{
    \begin{tabular}{|c||c|c|c|c|c|c|c|}\hline
        Work                                                                   & EE        & EE           & Decision   & MPSoC\tablefootnote{MPSoC: Multiprocessor System-on-Chip}     & DVFS\tablefootnote{DVFS: Dynamic Voltage and Frequency Scaling}   & Layer    \\
                                                                               & location  & architecture & Thresholds & Mapping   &        & Parallel  \\
        \hline \hline
        Optimal Location~\cite{chiangOptimalBranchLocation2021}                & \cmark    & \xmark       & \xmark     & \xmark    & \xmark & \xmark  \\\hline
        EExNAS~\cite{odemaEExNASEarlyExitNeural2021}                           & \cmark    & ?            & \xmark\tablefootnote{EExNAS calibrates the output confidence of the classifiers, but does not define a threshold value.}    & \xmark    & \xmark & \xmark  \\\hline
        HADAS~\cite{bouzidiHADASHardwareAwareDynamic2023}                      & \cmark    & \cmark\tablefootnote{HADAS utilizes the same \ac{EE} structure across all branches and backbone models.}       & \xmark     & \xmark    & \cmark & \xmark  \\\hline
        Map-and-Conquer~\cite{bouzidiMapandConquerEnergyEfficientMapping2023b} & \cmark    & ?            & \xmark     & \cmark    & \cmark & \cmark  \\\hline
        EDANAS~\cite{gambellaEDANASAdaptiveNeural2023a}                        & \cmark    & ?            & \xmark     & \xmark    & \xmark & \xmark  \\ \hline\hline
        Our \ac{NA} flow                                                       & \cmark    & \cmark       & \cmark     & \cmark    & \xmark & \xmark  \\ \hline
    \end{tabular}
    }
    \caption{A comparison of features within the state-of-the-art solutions for \ac{EENN}-\ac{NAS} frameworks and our \ac{NA} flow. Not all publications describe, if and how the \ac{EE} branches are configured.}
    \label{tab:compare_nas}
\end{table}

\section{Methodology}

In the previous research on \ac{NAS} for \acp{EENN}, multi-tiered evolutionary search has been the primary focus.
However, this approach is computationally expensive and resource-intensive.
To address this issue, we have developed a framework that is designed to reduce the search cost while still being able to find viable results.



The main input used by our framework is a pretrained base model.
In addition, training and validation sets are provided, along with a simple hardware description for each processor.
Lastly, the framework requires input on the order of processor usage, a description of the connections between processors, and the worst-case latency constraint.
Optionally, a weight parameter that balances efficiency gains and accuracy reduction penalties can be set.
During the search, the efficiency gain is defined as the share of the reduction in mean operations per inference compared to the original model.

The architecture search space is constructed from mutations of the base model.
It is pruned by evaluating only those that are predicted to fit within the given latency constraint or the memory sizes of the targeted devices.

We use worst-case latency as a constraint because it is critical for many embedded applications, especially in control loops, where delays in processing can have severe consequences.
Moreover, by considering worst-case latency, we ensure that the network can meet the latency requirements in all cases, not just on average.

We assume the \acp{EE} of an \ac{EENN} to be independent. 
This assumption is based on the similarity of \acp{EENN} to IDK classifier cascades~\cite{WangIDKCascades}.
These classifier cascades consist of multiple different models that are always queried in the same order and only proceed with the next classifier stage, if the current stage returns IDK (“I don’t know”), otherwise the inference is terminated.
For \acp{EENN} the decision mechanism is equivalent to the IDK label of these cascades.
While the \ac{EE} classifiers of an \ac{EENN} share weights through the feature extraction layers of the backbone, they do not operate on the same representation as they are attached at different locations of the backbone and the classifier that operates on the extracted features is distinct from the backbone model as it is its own branch without shared weights.
This enables the assumption that the predictions produced by the classifiers of an \ac{EENN} are not correlated.

This assumption allows us to individually evaluate the \acp{EE} and treat architectures within the search space as cascades of its contained classifiers.
This enables the flow to reuse the costly training steps across architectures but might limit their prediction performance without the joint-training step that is only applied to the found solution.

It also allows the \ac{NA} to efficienctsearch the optimal exit-wise threshold configuration for each architecture.
This is done by converting the confidence threshold search space into a directed graph.
The nodes represent the different threshold configurations of each exit.
The edges between the nodes are annotated with the impact on efficiency and accuracy when a specific threshold for an exit is combined with the thresholds of the other exits.
This turns the search for the optimal decision mechanism into a shortest-path problem, which can be solved rather quickly.

When searching for the best solution on the evaluated options, each architecture is only considered with its best decision configuration.
This drastically reduces the number of considered options during the final search step.
Once a solution has been found, an optional joint training step can be applied to the \ac{EENN}, which trains the model for one epoch to finetune the \acp{EE} and backbone.
If this optional step is applied, another search for the threshold configuration is performed afterward.

\subsection{Architecture Search}

The architecture search step is the process of finding the best architectures and locations for \acp{EE} that can be attached to the base model.
The first step of this process is translating the input model into two graph-level representations.
The fine-grained representation operates on a layer-level.
It is used to estimate the inference cost and derive a blueprint from the classifier of the base model.
The coarse-grained block-level representation collapses residual blocks and fuses post-processing layers with compute layers into nodes called blocks.
The block-level representation is used to identify possible locations for \acp{EE}.
The creation of these fused blocks reduces the number of locations that need to be evaluated without impacting the quality of the found architectures.
The nodes of both graphs are associated with their estimated cost in terms of \ac{MAC} operations, inference latency, memory, and storage footprint.
The current version of the framework uses simple approximations instead of accurate performance models to estimate the inference cost, as we focused on the search algorithm.



After identifying the possible locations for \acp{EE}, the \ac{EE} branches are configured for each location.
The architecture of each \ac{EE} is based on the classifier blueprint that was extracted from the backbone model.
Depending on the size of the \ac{IFM} at the location, additional downsampling layers might be inserted before the \ac{EE} classifier.
This process is rule-based and ensures that the added resource consumption of the \ac{EE} branches stays well below the overall inference cost of the model.
As the main focus is \ac{IoT} scenarios, the downsampling is applied aggressively to minimize the additional cost of the \ac{EE} branches.
To the best of our knowledge, our framework is the first to construct the \acp{EE} based on the original classifier, which ensures that they are able to perform the same task as it. 
It also enables the construction of \acp{EE} that are optimized for the required task and our framework can be extended to support tasks outside of the currently supported (binary) classification if a confidence metric can be defined for their results.

We limited the maximum number of classifiers that can be inserted to the number of target processors.
This limitation is due to the lack of support for conditional execution in most \ac{DL} frameworks and dedicated accelerators, especially in the embedded field.
Our framework adheres to these limitations to ensure compatibility with standard toolchains and simplifies the targeting of \ac{IoT} scenarios.
The alignment of \acp{EE} with the boundaries between processors is also intended to maximize the efficiency on platforms with separate power domains.
This stems from the observation, that deeper \acp{EE} tend to achieve higher accuracy scores and is intended to reduce the utilization of the following processors by maximizing the prediction quality and termination rate of the classifier assigned to the current processor.

\begin{figure}
    \centering
    \includegraphics[width=0.9\linewidth]{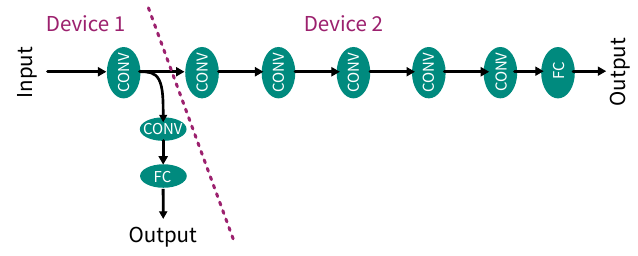}
    \caption{The \ac{EENN} architecture from Figure ~\ref{fig:eenn} has been mapped to a platform that contains two processing targets.}
    \label{fig:mapping}
\end{figure}

Side effects of these steps are the shrinking of the search space, a simplification of the deployed control flow, and a limitation of the overhead that can be introduced into the inference process.
The mapping of the possible solutions has been visualized in Figure~\ref{fig:mapping}, which illustrates the distribution of classifiers across the targeted processors.

The search process starts by generating a set of all \acp{EENN} versions of the base model that are estimated to lie within the given constraints and adhere to the maximum number of classifiers.
The assumed independence of \acp{EE} enables each \ac{EENN} to be evaluated as the weighted sum of the performance of the contained classifiers.
The used weights are the estimated termination rates of the classifiers.
Each evaluated \acp{EE} is trained and evaluated individually on the frozen base model.
Freezing the weights of the shared layers has the benefit of drastically reducing the training costs, as the \acp{EE} are very small compared to the backbone, enabling their training on consumer hardware.
Additionally, it prevents interactions between the classifiers during the training, maintaining their independence during this step, which is necessary to enable their reuse across evaluated options.
While related work like HADAS~\cite{bouzidiHADASHardwareAwareDynamic2023} also trains the \acp{EE} on frozen backbone weights during its inner optimization loop, it seemingly does not reuse their evaluation results across the different \ac{EENN} architectures within the search space and does not perform any joint training of the found solution.
These optimizations can greatly reduce the computational resources required and accelerate the search process.


Once all \acp{EE} of a possible solution have been evaluated, their threshold configuration is searched.
This is done for each evaluated \ac{EENN} architecture, as the final performance largely depends on the decision mechanism.
The relevant metrics of each possible architecture are predicted based on the performance of the contained classifiers and their termination thresholds.
These metrics include accuracy, early termination rates, mean inference time, and cost (in terms of \ac{MAC} operations).
By combining these metrics, we can associate each \ac{EENN} option with a scalar cost value that is the sum of the individual costs.


In the final step of the process, the evaluated \acp{EENN} are sorted by their cost, and the element with the lowest value is selected as the returned solution.
An optional fine-tuning step is intended to jointly optimize the weights of the shared feature extraction layers, as training the individual \acp{EE} on a frozen backbone can limit their achievable performance.
However, this step is optional to enable systems that do not have the compute capabilities to train the backbone model to still perform the \ac{NA}.

\subsection{Decision Mechanism Configuration}

The configuration of the decision mechanism for an \ac{EENN} is important for the final performance of the model, as it significantly impacts accuracy, latency, and energy efficiency. 

The search is performed by evaluating the relevant metrics on the validation set for a discretized range of possible thresholds on each \ac{EE}.

The validation set or a similar set of unseen samples is used to get confidence values of the \acp{EE} for unseen samples.
If no such set is available, our framework will fall back to the training set.
This fallback will result in overly optimistic decision configurations due to the higher confidence on samples that the model was trained on.
This is compensated by scaling the found thresholds with a correction factor.

\begin{figure}
    \centering
    \includegraphics[width=0.9\linewidth]{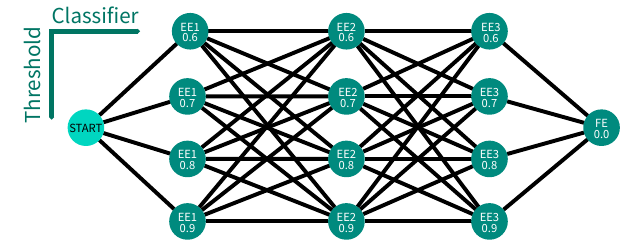}
    \caption{An illustration of the graph structure that is used for the search of exit-wise confidence thresholds for an \ac{EENN} with three \acp{EE}.
    This illustration has limited the range of evaluated thresholds for readability purposes.}
    \label{fig:decision_search_space}
\end{figure}

A search graph is constructed based on the evaluated values, where each node represents a tuple of the classifier and its associated threshold.
Such a search graph has been visualized in Fig.~\ref{fig:decision_search_space}.
The edges represent the change in relevant metrics between these nodes and are associated with a scalar cost value during the search.
We use the Bellman-Ford shortest path algorithm to search for the optimal decision mechanism, as edges do not necessarily have to be associated with a positive value.
However, the size of the search graph is small enough that the difference in cost compared to Dijkstra is negligible.


The search graph is limited in size as thirteen nodes are inserted per early classifier that is part of the \ac{EENN} architecture.
For an \ac{EENN} with two \acp{EE} and one final classifier, the resulting search graph for an architecture option has 28 nodes:
two \acp{EE}, one input node, and one node for the final classifier that is associated with a threshold of zero, as all remaining samples will terminate there.


The optional second search step for the selected \ac{EENN} allows for the consideration of significantly more thresholds.



\section{Evaluation}

The evaluation focuses on assessing the performance of \acp{EENN} in typical embedded \ac{DL} scenarios.
To this end, we convert and deploy a speech command recognition and \ac{ECG} classification model to a heterogeneous \ac{MCU}.

The Infineon PSoC~6\textsuperscript{\tiny\textregistered} \ac{MCU} was selected, as it contains an ARM Cortex-M0Plus\textsuperscript{\tiny\textregistered} (M0) and an additional Cortex-M4F\textsuperscript{\tiny\textregistered} (M4F) processor that share up to 1MB of SRAM memory and up to 2MB of Flash storage.
The M0 core operates at 100 MHz, while the M4F operates at 150 MHz when active.
Despite its multi-core architecture, the memory is single-ported, which prevents the cores from running concurrently.
Only one core can be utilized at a time, leaving the other processor in a sleep state.

The energy values reported are estimates based on the measured runtimes and the reported power consumption for the associated states of the \ac{MCU} according to the technical documentation of the platform\footnote{PSoC6 CY8C624ABZI-D44 data sheet: \url{https://www.infineon.com/dgdl/Infineon-PSOC_6_MCU_CY8C62X8_CY8C62XA-DataSheet-v16_00-EN.pdf?fileId=8ac78c8c7d0d8da4017d0ee7d03a70b1}}.

We propose mapping \acp{EENN} to the \ac{MCU} by using the M0 core as an always-on device that performs permanent monitoring within a small energy envelope.
In cases where uncertainty arises, the M4F core will be used to provide additional prediction capabilities.
This allows for high accuracy in permanent monitoring while maintaining high energy efficiency.
This approach is expected to result in extended battery life, making it suitable for scenarios where this is a critical factor.

All searches were performed on an Intel Core i5-1240P mobile CPU with 64\,GiB of system memory, TensorFlow 2.13 for all training and prediction steps, and without GPU acceleration.
An overview of the experimental results can be found in Tab.~\ref{tab:results}.

\begin{table}
    \centering
    \resizebox{\textwidth}{!}{
    \begin{tabularx}{1.4\textwidth}{| C{1.5cm} || Y | Y || Y | Y | Y | Y || Y | Y | Y | Y |}
    \hline
        Model       & ARM-DS (L)   & 1D-CNN             & \multicolumn{8}{c|}{ResNet-152} \\ \hline
        Calibration & val.         & val.               & 1            & 2/3                & 1/2               & val.          & 1             & 2/3               & 1/2           & val.          \\ \hline
        Dataset     & GSC          & MIT-BIH            & \multicolumn{4}{c|}{CIFAR-10}                                         & \multicolumn{4}{c|}{CIFAR-100}                                    \\ \hline
        Target      & \multicolumn{2}{c||}{PSoC6:}      & \multicolumn{8}{c|}{Cortex-A55+A76,}                                                                                                       \\
                    & \multicolumn{2}{c||}{Cortex-M0P,} & \multicolumn{8}{c|}{Mali G610,}                                                                                                            \\
                    & \multicolumn{2}{c||}{Cortex-M4F}  & \multicolumn{8}{c|}{LTE Uplink (50Mbps),}                                                                                                  \\
                    & \multicolumn{2}{c||}{}            & \multicolumn{8}{c|}{RTX\,3090\,Ti}                                                                                                         \\ \hline \hline
        Training    & 89\,min       & 17\,min           & \multicolumn{3}{c|}{pretrained}                       & 1,1119\,min   & \multicolumn{3}{c|}{pretrained}                   & 1,076\,min    \\ \hline
        Search      & 18\,min       & 7\,min            & \multicolumn{3}{c|}{504\,min}                         & 270\,min      & \multicolumn{3}{c|}{561\,min}                     & 266\,min      \\ \hline \hline
        Acc.        & 84.45\%       & 96.5\%            & 92.81\%       & 86\%              & 72.72\%           & 92.66\%       & 72.56\%       & 72.49\%           & 71.85\%       & 70.48\%       \\
                    & \tbf{-12.96}  &\tbf{-3.1}         & \tbf{-1.18}   & \tbf{-7.99}       & \tbf{-21.25}      & \tbf{-0.32}   & \tbf{+0.02}   & \tbf{-0.05}       & \tbf{-0.69}   & \tbf{+0.65}   \\ \hline
        Prec.       & 91.6\%        & 99.4\%            & 92.84\%       & 86.2\%            & 73\%              & 92.68\%       & 73.04\%       & 72.92\%           & 72.29\%       & 70.72\%       \\
                    & \tbf{-7.77}   &\tbf{-0.2}         & \tbf{-1.25}   & \tbf{-8.09}       & \tbf{-21.08}      & \tbf{-0.31}   & \tbf{-2.31}   & \tbf{-2.43}       & \tbf{-3.06}   & \tbf{+0.52}   \\ \hline
        Recall      & 84.4\%        &98.6\%             & 92.81\%       & 86\%              & 72.74\%           & 92.66\%       & 72.56\%       & 72.49\%           & 71.85\%       & 70.48\%       \\ 
                    & \tbf{-14.93}  &\tbf{-1.0}         & \tbf{-1.11}   & \tbf{-7.89}       & \tbf{21.15}       & \tbf{-0.32}   & \tbf{+1.11}   & \tbf{+1.04}       & \tbf{+0.4}    & \tbf{+0.65}   \\ \hline \hline
        Mean        & 11.8M         &0.33M              & 318.12M       & 226.04M           & 147.99M           & 330.97M       & 357.24M       & 349.4M            & 342.73M       & 358.28M       \\
        MACs        & \tbf{-59.67\%}&\tbf{-78.3\%}      & \tbf{-11.3\%} & \tbf{-36.99\%}    & \tbf{-58.75\%}    & \tbf{-7.75\%} & \tbf{-0.43\%} & \tbf{-2.61\%}     & \tbf{-4.47\%} & \tbf{-0.13\%} \\ \hline
        Mean        & 1.06\,sec     & 0.62\,sec         & 16.2\,ms      & 12.35\,ms         & 8.6\,ms           & 16.28\,ms     & 21.05\,ms     & 19.57\,ms         & 18.94\,ms     & 17.19\,ms     \\
        Latency     & \tbf{+34.37\%}&\tbf{-58.9\%}      & \tbf{-9.18\%} & \tbf{-30.79\%}    & \tbf{-51.79\%}    & \tbf{-8.77\%} & \tbf{+19.72\%}& \tbf{+11.33\%}    & \tbf{+7.76\%} & \tbf{-2.2\%}  \\ \hline
        Mean        & 21.3\,mJ      &11.83\,mJ          & -             & -                 & -                 & -             & -             & -                 & -             & -             \\
        Energy      & \tbf{-13.6\%} &\tbf{-74.9\%}      &               &                   &                   &               &               &                   &               &               \\ \hline
        Early Term. & 83.4\%        & 100\%             & 36.99\%       & 86.97\%           & 95.4\%            & 31.16\%       & 13.69\%       & 61.65\%           & 74.39\%       & 0.33\%        \\ \hline
    \end{tabularx}
    }
    \caption{The performance of the created \acp{EENN} compared to placing the entire original network on a single processor of the target platform (either the M4F core or the Mali GPU).
    The difference in relevant performance metrics to the submitted model are marked in bold.}
    \label{tab:results}
\end{table}

\begin{figure}
    \centering
    \includegraphics[width=1\linewidth]{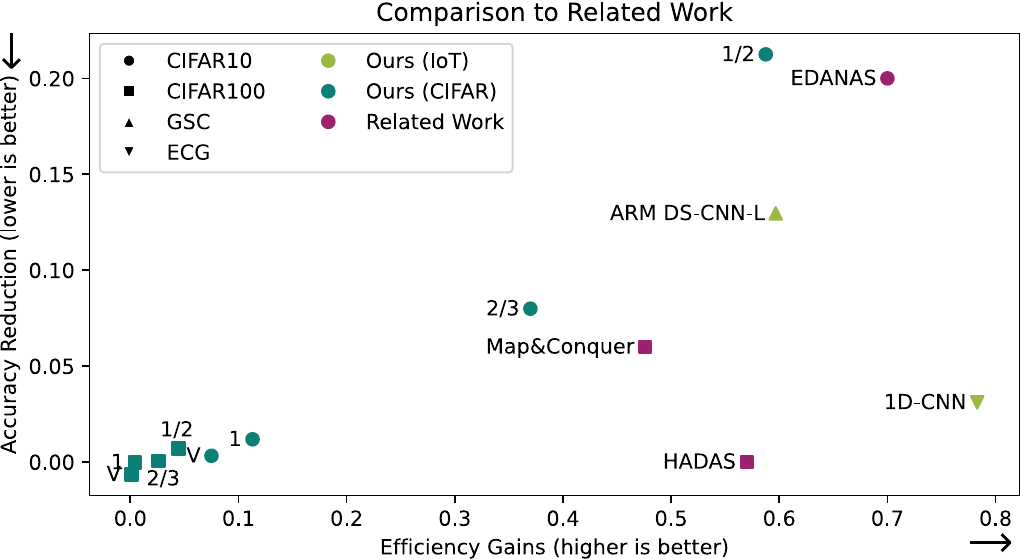}
    \caption{
    Performance comparison of our framework to previous work using different base models on the same datasets.
    }
    \label{fig:scatter}
\end{figure}



\subsection{Speech Command Detection}

The first experiment was the detection of speech commands on audio data.
The Google Speech Command dataset~\cite{wardenSpeechCommandsDataset2018a} was used in combination with a depthwise-separable CNN from ARM in its largest version~\cite{zhangHelloEdgeKeyword2018a}.
The model performs a classification into eleven different classes: nine commands, silence, and background noise, mostly consisting of spoken words that are not part of the set of commands.
Possible applications for this are either wake word detection or accessibility features.

The reference model achieves 99.33\% accuracy, 99.37\% precision, and 99.33\% recall.
The search was configured with a desired worst-case latency of 2.5 seconds.
The required hardware description specifies the estimated processing speed of both cores in MAC operations per second, their interconnect speed, and the available memory size.
The interconnect speed was defined as the theoretical speed of the used memory.
The processing speeds of the cores were estimated at 10 million MAC operations per second for the M0 and 75 million for the M4F.
The search was parameterized with a weight of 0.9 on reduction of the inference cost and 0.1 on maintaining accuracy.

The search space consists of six possible architectures, one of which was rejected before its accuracy evaluation due to being estimated to be outside of the latency constraint.

The selected solution adds an \ac{EE} after the second convolutional layer of the backbone and configures its confidence threshold to 0.6.
The \ac{EE} on its own achieved 74.99\% accuracy, 86.34\% precision, and 74.93\% recall.
The inference of the entire \ac{EENN} takes 967.99\,ms for the M0 subgraph and an additional 521\,ms for the second subgraph that has been mapped to the M4F.
This results in a worst-case latency of 1.5 seconds, which lies well within the set constraint.

The execution of the M0 workload requires 18.53\,mJ of energy, and the M4F subgraph requires an additional 16.65\,mJ if it needs to be executed.

Despite the significant decrease in mean \ac{MAC} operations per inference, the M0's capabilities limit the achievable efficiency gains.
In contrast to the M4F, there is no dedicated instruction or unit to efficiently perform \ac{MAC} operations on the M0 core.
As a result, the performance improvements are modest in comparison to the reduction in MAC operations.
Due to the promising reduction in MAC operations, we remain optimistic about the use with future generations of \acp{MCU} that feature more advanced cores within the always-on power domain.
\subsection{ECG Classification on Wearable Devices}

The second experiment targets the classification of single-lead ECG signals using a convolutional \ac{NN}.
The MIT-BIH dataset~\cite{moodyMITBIHArrhythmiaDatabase1992} was used and a fully convolutional \ac{NN} was utilized as the backbone model~\cite{HeartbeatClassificationBased2023}.
The data was split into dedicated train, validation, and test sets.
The model classifies the \ac{ECG} signals into six classes:
normal heartbeat, atrial premature beat, premature ventricular contractions, right-bundle or left-bundle branch block beat, and paced beat.
Premature beats and block beats are indicators of underlying health conditions that medical experts should investigate.
The model achieved 99.29\% accuracy, 99.33\% precision, and 99.25\% recall on the test set.
A possible application is the deployment to wearable devices like smartwatches to enable constant monitoring. 

The previous experiment's hardware description and search configuration were adapted to account for additional reshaping operations inserted by the embedded \ac{DL} toolchain (TensorFlow Lite for microcontrollers).

The resulting \ac{EENN} inserted the \ac{EE} after the first convolutional block and configured its threshold to 0.6.
The execution of the first subgraph takes 618\,ms on the M0, and the second subgraph 1.376\,seconds on the M4F.

The high termination rate indicates that the model might be overparameterized.
However, the option to execute the entire \ac{EENN} is beneficial for healthcare applications as it ensures a high accuracy of the classification results if needed.

\subsection{Image Classification}

To ensure that our results can be compared to established work, we conducted experiments using the CIFAR-10 and -100 datasets~\cite{Krizhevsky09learningmultiple} and converted ResNet-152~\cite{Resnet} accordingly.
We used a distributed system consisting of a Rockchip RK3588\textsuperscript{\tiny\textregistered}-based platform, which includes four Cortex-A76 and four Cortex-A55 cores, along with a Mali G610 GPU and 16\,GiB of shared system memory.
The CPU cores are grouped into a single target for our framework so as not to interfere with the operating system's scheduler and optimized kernels for this platform.
Additionally, the local device has a 50\,Mbps connection to a workstation with an Nvidia GeForce RTX\,3090\,Ti to simulate a smartphone connected to a cloud backend.

We opted to use ResNet-152 as a particularly challenging test case.
This model's depth creates 74 potential \ac{EE} locations, resulting in a total of 2,776 possible \ac{EENN} architectures for our target system.
Each of these architectures has up to 169 threshold configuration options, resulting in a search space of approximately 450,000 configurations.
We found that reducing the latency constraint did not significantly decrease the overall search time.
Even with fewer architectures to evaluate, the majority of the search time is spent on training and evaluating all potential \acp{EE}, which are still contained in the remaining architectures.

We performed two evaluations of the conversion process: one with a dedicated calibration/validation set (20\% of the training set) and another without it.
In the latter, we used pretrained models~\cite{ResNet_pretrained} and calibrated the thresholds using the training data.
A correction factor is applied after calibration to account for the increased confidence compared to unseen samples.
Different correction factors have been evaluated.

Our results indicate that using a correction factor and only the training set can be a viable alternative to a calibration set, particularly when the latter is unavailable.
The solutions that used a dedicated calibration set achieved the lowest reduction in prediction quality and the fastest search times.
The reduction in search time was achieved as the framework uses the calibration set to terminate the evaluation of \acp{EE} early.
This early termination uses the calibration set performance after the first training epoch of the \ac{EE} to decide if the classifier is able to achieve a meaningful prediction quality.
On the other hand, the \acp{EENN} models that were created using only a training set and a correction factor achieved significantly higher efficiency gains and local termination rates but showed a larger decrease in prediction quality.
This result is caused by generating threshold configurations that are lower than the lower boundary of the search space by applying the correctionf factor.
This showcases that some of the design decisions within our framework that are intended to target low-energy \ac{IoT} applications with smaller network sizes and a limited number of classes interfere with the conversion of larger computer vision models like ResNet-152 and the 100 classes of CIFAR-100.

Despite these limitations our framework was able to discover solutions for this larger problem sizes using consumer-grade hardware in a short time frame.
Additionally, all found solutions achieved (limited) efficiency gains. 
While the created \acp{EENN} did not always show the same level of improvement as established work that also utilized additional optimizations (see Tab.~\ref{tab:compare_nas}), as depicted in Fig.~\ref{fig:scatter}, they still achieved usable results. 

The design decisions that limit the performance include the rule-based system to adapt the \ac{EE} classifiers to the backbone, which aims to minimize the overhead.
This system was designed for \ac{IoT} applications with shallower models and smaller \acp{IFM} and resulted in \acp{EENN} whose branches total up to a footprint of less than 0.5\% of the backbone's \ac{MAC} operations.
This design choice reduces the predictive capabilities of \ac{EE} classifiers in the first half of \ac{EE} locations, which limits their potential contribution.
This limited the performance of the solutions within the search space.
More research is needed to extend the rule-based system to larger \ac{IFM} and network sizes.
Additionally, the threshold search space was designed for the lower class count of most embedded use cases, which limited the quality of its results on the CIFAR-100 dataset.

As shown in Table~\ref{tab:results}, our approach requires up to 9.4 hours of search time for this problem on the used mobile CPU to analyze the search space of 2,776 possible \ac{EENN} architectures and their optimal decision mechanism configuration.
Related work relied on clusters of twelve to 32 unnamed GPUs, accumulating between twelve hours and three days of search time to find an \ac{EENN} without its optimal decision mechanism~\cite{bouzidiMapandConquerEnergyEfficientMapping2023b,bouzidiHADASHardwareAwareDynamic2023}.
The CPU used for the evaluation of our \ac{NA} flow has a peak power consumption of less than 70\,W, which is significantly lower than the sustained power consumption of data center GPUs used by previous publications.
A direct comparison between our \ac{NA} and the previous work is not meaningful due to the difference in functionality and features.
However, the exhaustive training and evaluation of the 2,776 \ac{EENN} architectures would require about 86.75 days of compute time on the used CPU.
This duration was estimated based on the duration of the single fine-tuning epoch applied to the found solution (540~seconds for one epoch), the assumption that five epochs would be necessary for each architecture in the search space and 2,776 contained architectures.
This time does not include the steps necessary to configure the decision mechanism for each option or the selection of the solution from the search space.


\section{Conclusion}

Our framework has demonstrated the ability to augment and map submitted models for the use as \acp{EENN} on distributed or heterogeneous devices. 
By modifying submitted standard models, it allows developers to quickly deploy their models to heterogeneous and distributed \ac{IoT} environments without the need for dedicated supernets or expert knowledge on adaptive inference.
It can often find \ac{EENN} versions of the backbone model in a shorter time frame than the training of the backbone itself requires and also fully configures the required decision mechanism within this time window.

We have applied the framework to a range of applications and target devices and have shown that it can identify viable solutions that reduce the mean inference cost, even for workloads it was not designed for.
To the best of our knowledge, our \ac{NA} framework is the first to cover the conversion into an \ac{EENN} architecture, its mapping to a heterogeneous platform and the configuration of the optimal confidence-based decision mechanism in a fully automatic process.

Moving forward, there are several areas that could be explored to improve our framework.
One possibility is to enhance the search algorithm to further increase the quality of the found solutions.
Additionally, the framework could be adapted to better support the conversion of larger \acp{NN} beyond the \ac{IoT} domain.
Lastly, the training and evaluation of the \ac{EE} classifiers is a bottleneck in the search process. 
Future work could explore the option of predicting the possible performance of an \ac{EE} based on the backbone model, the attachment location, and the \ac{EE} architecture. 
This approach could significantly reduce search time and be used to guide the search towards options that are predicted to be of higher quality.

\section*{Acknowledgement}
The project “RadarSkin” has received funding from the \ac{BMBF} under the call “Electronic Systems for Edge Computing” (grant number 16ME0543). The responsibility for the content of this publication lies with the author.

\bibliographystyle{splncs04}
\bibliography{Audio,ECG,EENN,NAS,RESNET}

\end{document}